\documentclass[sigconf]{acmart}

\usepackage{multirow}
\usepackage[noend]{algpseudocode}
\usepackage{algorithmicx,algorithm}
\usepackage{wrapfig,lipsum,booktabs}
\usepackage{algorithmicx,algorithm}

\newcommand{\z}{{\rm\bf z}}                   
\newcommand{\w}{{\rm\bf w}}                   
\newcommand{\txt}{{\rm\bf t}}                   
\newcommand{\Z}{\mathcal{Z}}                  
\newcommand{\W}{\mathcal{W}}                  
\newcommand{\x}{{\rm\bf x}}                   
\newcommand{\X}{\mathcal{X}}                  
\newcommand{\Loss}{\mathcal{L}}               

\AtBeginDocument{%
  \providecommand\BibTeX{{%
    \normalfont B\kern-0.5em{\scshape i\kern-0.25em b}\kern-0.8em\TeX}}}

\copyrightyear{2022}
\acmYear{2022}
\setcopyright{acmlicensed}\acmConference[MM '22]{Proceedings of the 30th ACM International Conference on Multimedia}{October 10--14, 2022}{Lisboa, Portugal}
\acmBooktitle{Proceedings of the 30th ACM International Conference on Multimedia (MM '22), October 10--14, 2022, Lisboa, Portugal}
\acmPrice{15.00}
\acmDOI{10.1145/3503161.3547809}
\acmISBN{978-1-4503-9203-7/22/10}

\begin{document}

\title{Paired Cross-Modal Data Augmentation for Fine-Grained Image-to-Text Retrieval}

\author{Hao Wang}
\affiliation{%
  \institution{Nanyang Technological University}
  \country{Singapore}
}
\email{hao005@ntu.edu.sg}

\author{Guosheng Lin}
\affiliation{%
  \institution{Nanyang Technological University}
  \country{Singapore}
}
\authornote{Corresponding authors}
\email{gslin@ntu.edu.sg}

\author{Steven C. H. Hoi}
\affiliation{%
  \institution{Singapore Management University}
  \institution{Salesforce Research Asia}
  \country{Singapore}
}
\email{chhoi@smu.edu.sg}

\author{Chunyan Miao}
\affiliation{%
  \institution{Nanyang Technological University}
  \country{Singapore}
}
\authornotemark[1]
\email{ascymiao@ntu.edu.sg}

\renewcommand{\shortauthors}{Hao Wang, Guosheng Lin, Steven C. H. Hoi, \& Chunyan Miao}

\begin{abstract}
This paper investigates an open research problem of generating text-image pairs to improve the training of fine-grained image-to-text cross-modal retrieval task, and proposes a novel framework for paired data augmentation by uncovering the hidden semantic information of StyleGAN2 model. Specifically, we first train a StyleGAN2 model on the given dataset. We then project the real images back to the latent space of StyleGAN2 to obtain the latent codes. To make the generated images manipulatable, we further introduce a latent space alignment module to learn the alignment between StyleGAN2 latent codes and the corresponding textual caption features. When we do online paired data augmentation, we first generate augmented text through random token replacement, then pass the augmented text into the latent space alignment module to output the latent codes, which are finally fed to StyleGAN2 to generate the augmented images. We evaluate the efficacy of our augmented data approach on two public cross-modal retrieval datasets, in which the promising experimental results demonstrate the augmented text-image pair data can be trained together with the original data to boost the image-to-text cross-modal retrieval performance. 
\end{abstract}



\maketitle

\section{Introduction}

It is often difficult or expensive to collect large amounts of data annotations for deep learning based model training, thus automated data augmentation has been widely used as a practical technique to boost the model performance. Existing methods \cite{guo2019augmenting,yang2021free,salvador2021revamping} are mainly restricted within the single-modality data augmentation. Specifically, Rand-Augment \cite{cubuk2020randaugment} and RandomErasing \cite{zhong2020random} are adopted on the vision transformers \cite{touvron2021training} to boost the model performance. In the cross-modal domain of image-text data augmentation, Mahajan et al.\cite{mahajan2020diverse} replace the object tokens in the captions with a placeholder to describe the context of the image, then they generate pseudo captions with similar semantic meanings to supervise captioning model learning. Wang et al. \cite{wang2018learning} use regions of high overlap with the ground truth regions as the augmented positive regions, such that the augmented region-phrase pairs give better matching results at evaluation. However, these prevailing data augmentation methods fail to change the semantic contents of the original data and cannot generate data with large diversity. Moreover, there are few works on the data augmentation techniques on the paired cross-modal text-image data. If we want to do the data augmentation for text and image both and construct useful text-image pairs, the challenge appears to be: how can we generate the augmented text-image pairs with the same semantic information?

To tackle the problem of paired data augmentation for the image-to-text cross-modal retrieval task, we propose a novel paired text-image data augmentation algorithm, which can be used together with the prevailing single-modality data augmentation strategy and is easily plugged into the existing retrieval methods. To be specific, since text is formed by the combinations of various word tokens, increasing the number of semantic word combinations can be a simple yet effective way to produce more diverse text features for the robust inference. That means we can randomly replace part of the words in the textual captions, to construct the augmented text. To further match the semantic consistency between the augmented text and images, we generate the augmented images from the augmented text. 

\begin{figure*}
\begin{center}
\includegraphics[width=0.82\textwidth]{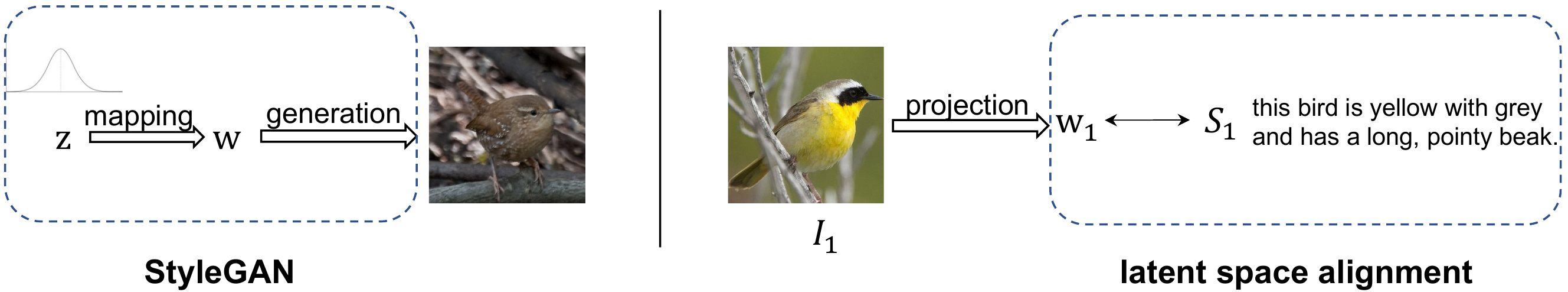}
\vspace{-15pt}
\end{center}
   \caption{The demonstration of the latent space alignment module of our algorithm. After the training of StyleGAN2, we project the real images back and get the latent codes $\w$ that can be used to reconstruct the given images. To manipulate the generated images based on the textual captions, we align the text features to the paired $\w$. With the learned latent space alignment module, we can generate images with the text semantic information, by giving the corresponding latent codes $\w$.
}
\label{fig:demo}
\vspace{-5pt}
\end{figure*}

We take the advantage of the StyleGAN2 \cite{Karras2019stylegan2} model, which can generate images with high quality and diversity. It is notable that we train the StyleGAN2 without the conditional text input, otherwise the model generation performance would be restricted by the limited text-image pairs \cite{wang2021cycle}. Since the latent space $\W$ of StyleGAN has been proven to be disentangled to the semantic contents \cite{shen2020interpreting,karras2019style}, the disentanglability of StyleGAN2 enables the effective semantic manipulation on the generated images. To this end, we first project the real images back to the latent space $\W$ of the trained StyleGAN2, where we can obtain the latent codes $\w$ that can be used to reconstruct the given images \cite{karras2020training}. With the projected $\w$ and the corresponding text captions, we map the text features to the space $\W$ and learn the alignment model between paired $\w$-text feature representations. The procedure is presented in Figure \ref{fig:demo}. We feed the augmented text into the trained latent space alignment module, and the output can be used as the latent codes $\w$ for StyleGAN2 to generate the augmented images. Thus we can get the paired augmented text-image data with consistent semantics.

To evaluate the efficacy of the augmented paired data, we experiment on the fine-grained image-to-text cross-modal retrieval task. Limited by the generation capacity of existing generative models, we can hardly generate images with multiple objects, such as images of COCO dataset \cite{lin2014microsoft}. The recently proposed XMC-GAN \cite{zhang2021cross} adopts complex architecture to produce decent COCO images, which is infeasible to online generate the augmented images from text during the retrieval training. Therefore, here we focus on the datasets with single-object images.

In the image-to-text cross-modal retrieval task \cite{li2020unicoder,liu2020graph,salvador2017learning}, given a sample from one modality (e.g.,  text), the model is required to find the corresponding data samples from another modality (e.g., image), or vice versa. Since our cross-modal data augmentation method can give unlimited raw text-image pairs in an online fashion, it can be used on top of the existing data augmentation methods and retrieval models. We experiment with multiple settings and model backbones, the experiments demonstrate models using our method can boost the original performance on two public datasets. Finally, we also present qualitative results of the augmented data.

\textbf{Our contributions.} In this paper, we develope a novel framework to generate new text-image data pairs to address the problem of cross-modal data augmentation. Our model consists of several novel contributions: (i) we present a method to tackle the challenge of semantic consistency between the generated text and images, which can be achieved through the projected latent codes of the StyleGAN2 model; (ii) we construct the augmented text by random token replacement, then pass the augmented text into the latent space alignment module to give the latent codes, which are fed into StyleGAN2 to generate the augmented images;  
and (iii) we adapt the proposed algorithm to the image-to-text retrieval task and boost the benchmark model performance.

\section{Related Work}
\textbf{Data augmentation.} 
For image augmentation, random rotation and flipping are some classic methods to obtain highly generalized deep networks \cite{simonyan2014very,he2016deep,huang2019convolutional}. To increase the robustness to adversarial examples, Zhang et al. propose mixup \cite{zhang2017mixup} to use the convex combinations of pairs of examples for model training, it can also benefit some downstream tasks such as the long-tailed recognition \cite{zhong2021improving}. Other than giving new samples to increase the data diversity, it is shown that abandoning certain information of the training images can also improve the model performance, for instance Cutout \cite{devries2017improved} randomly cuts a region out the original image for augmentation. OnlineAugment \cite{tang2020onlineaugment} can adapt online to the learner state throughout the entire training and learn a wide variety of local and global geometric and photometric transformations.
For text augmentation, there are also mixup methods. Guo et al. \cite{guo2019augmenting} combine the word or sentence embeddings in a certain portion to perform the augmentation on the text embeddings. Training with adversarial examples \cite{belinkov2017synthetic,zhao2017generating,zhu2019freelb} can also be used to improve the generalization of language models. Belinkov et al. \cite{belinkov2017synthetic} manipulate each word in the sentences with natural noise or synthetic and further increase model robustness. Zhuang et al. \cite{zhuang2019fmri} use generative models to improve the fMRI classification performance. Wang et al. \cite{wang2021pointaugmenting} augment the LiDAR Points and camera images for the 3D object detection task. 

\begin{figure*}
\begin{center}
\includegraphics[width=0.8\textwidth]{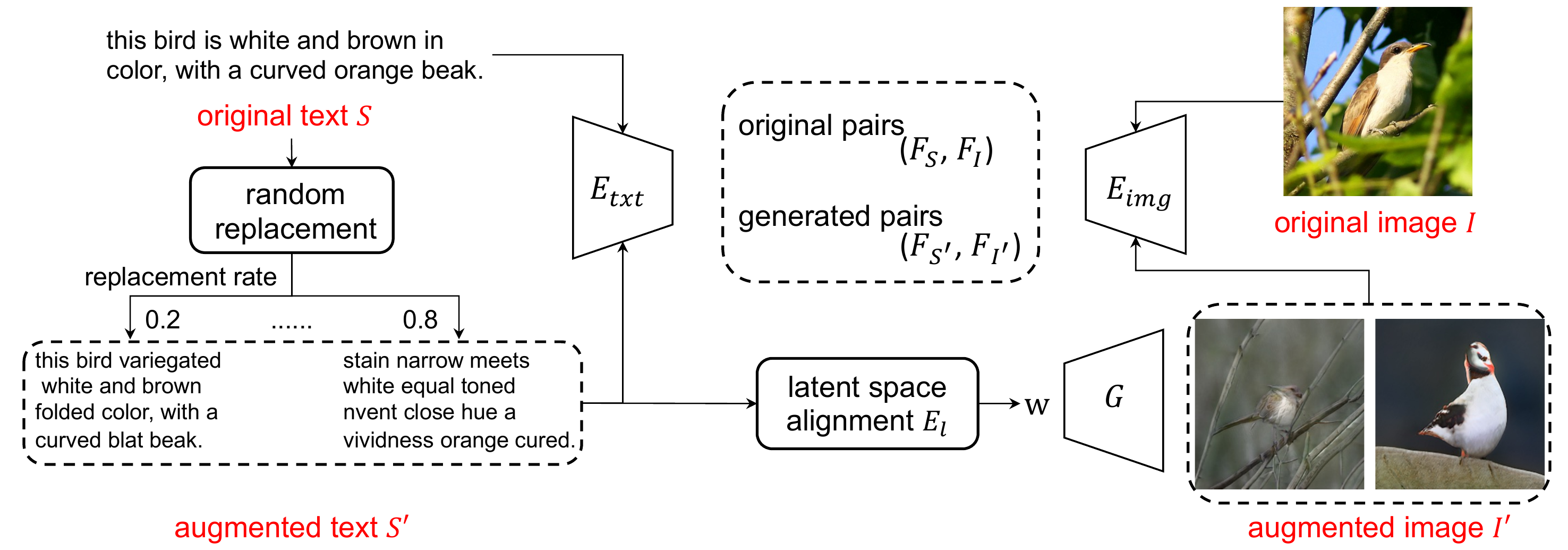}
\end{center}
\vspace{-15pt}
   \caption{The generation flow of our proposed paired cross-modal data augmentation. We first replace part of the text tokens randomly, which can be controlled by different replacement rates. To further generate the augmented images based on the augmented text, we learn a latent space alignment module $E_l$. In this module, we align the StyleGAN2 latent codes $\w$ with the paired text features. The trained alignment module $E_l$ is used to interpret the augmented text semantic information as $\w$, which is fed into the StyleGAN2 to produce the augmented images. $E_{txt}(\cdot)$ and $E_{img}(\cdot)$ are the text and image encoders respectively. We incorporate both the original data representations $(F_S, F_I)$ and the online generated text-image data representations $(F_{S'}, F_{I'})$ into the retrieval training. 
}
\label{fig:framework}
\end{figure*}

\textbf{Cross-modal retrieval.} The cross-modal retrieval task aims to retrieve the corresponding instance of different modalities based on the given query. The general idea is to map the cross-modal data to a common space, such that the heterogeneous data can be correlated. To capture the complex non-linear correlations between text-image data, most existing methods \cite{li2020weakly,wang2019learning,liu2020graph,jing2020incomplete,li2020unicoder} are based on the deep model architecture. Specifically, to learn the fine-grained correspondence across different semantic levels, Liu et al. \cite{liu2020graph} construct textual and visual graphs, where they achieve the cross-modal correlation through the node-level and structure-level matching. PVSE \cite{song2019polysemous} learns the one-to-many mappings, which produces multiple candidate samples for a given input. PCME \cite{chun2021probabilistic} uses the probabilistic model while PVSE uses deterministic model, PCME yields better performance than PVSE.
To make use of the unpaired data in the given dataset, Jing et al. \cite{jing2020incomplete} adopt autoencoders to obtain the modality-invariant representations, which can be achieved by the dual alignment at the distribution and the semantic levels. The triplet loss is widely used \cite{liu2020graph,wang2019learning,lee2018stacked} as the objective for retrieval training.

\section{Method}
Our proposed paired cross-modal data augmentation method is shown in Figure \ref{fig:framework}. The whole pipeline can be summarized as the following three-stage training scheme:
\begin{itemize}
   \item \textbf{Stage 1.} We train a StyleGAN2 \cite{Karras2019stylegan2} model with the images only. The StyleGAN2 model maps the random noise space $\Z$ to the style latent space $\W$, which is disentangled and helps generate images with high quality and diversity.  
   \item \textbf{Stage 2.} We project the real images back to the latent space $\W$ and obtain the latent codes $\w$ for given images. We then learn a latent space alignment module $E_l$, where the text features are mapped to align with the corresponding latent codes $\w$.
   \item \textbf{Stage 3.} We do the cross-modal data augmentation in an online fashion. The augmented text is constructed by random token replacement. Then we feed the augmented text into the trained alignment module $E_l$, whose output can be used as the latent codes $\w$ for StyleGAN2 to generate the augmented images. 
\end{itemize}

We present more technical details for each stage in the following sections.

\subsection{Image projection to latent space}
The StyleGAN2 model can be denoted as $G(\cdot): \Z\rightarrow\X$, where the model uses Multi-Layer Perceptron (MLP) to map the initial noise space $\Z$ to the style latent space $\W$. Then the StyleGAN2 generates images based on latent codes $\w$ of the disentangled space $\W$.
After the training of StyleGAN2 model on the given datasets, we follow \cite{Karras2019stylegan2} to project the real images back to the latent space $\W$. 

In this module, we take the latent codes $\w \in \W$ for optimization. Specifically, we first run 10,000 random noise input $\z$ to produce mapped latent codes $\w = \mathrm{MLP}(\z)$. We use the average $\mu_{\w} = \mathbb{E}_{\z}\mathrm{MLP}(\z)$ as the initialization of $\w$, and the approximated scale of $\W$ can be set as $\sigma^2_{\w} = \mathbb{E}_{\z}||\mathrm{MLP}(\z)-\mu_{\w}||_2^2$, which is the average square Euclidean distance to the center. We adopt $\tilde{\w} = \w + \mathcal{N}(0, 0.05\sigma_{\w}k^2)$ as the input to generate images, where $k$ goes from one to zero gradually.
This is an empirical formulation used in \cite{Karras2019stylegan2}. The usage of Gaussian noise on $\w$ adds stochasticity to the optimization process and makes the finding of the global optimum stabilized.

Our goal is to make the reconstructed images from the projected latent codes $\x'=G(\tilde{\w})$ identical to the original real images $\x$. To this end, we adopt the perceptual loss \cite{johnson2016perceptual} as the optimization objective, which can be denoted as 
\begin{align}
  \begin{aligned} \label{eq:pro}
    \min_{\w} \Loss_{p} = ||F(\x) - F(G(\tilde{\w}))||^2_2,
  \end{aligned}
\end{align}
where the $F(\cdot)$ denotes the VGG \cite{simonyan2014very} feature extraction model. The optimized results $\w_{opt}$ are the closest latent codes that can reconstruct the given real images.

\subsection{Latent space alignment}
The projection of images to latent space $\W$ gives the corresponding latent codes $\w_{opt}$ for images $\x$. As the latent space $\W$ of StyleGAN2 has been proven to be disentangled to the semantic contents \cite{shen2020interpreting,karras2019style}, which has the same property as text feature space. We can map the text representations into the same space as $\W$. Moreover, since the text representations can be semantically changed with the raw text input, when we can achieve the multi-modal alignment between text feature space and StyleGAN2 latent space $\W$, the generated images can be manipulated with the given text. 

To be specific, we adopt an LSTM $E_l$ to encode the textual captions $S$ and output the text representations $\txt=E_l(S)$, where $\txt$ has the same feature dimensions as $\w$. Since the paired relationships between images and latent codes as well as that between images and text are available, we can adopt the pairwise ranking loss to learn the alignment between $\txt$ and $\w_{opt}$, which can be trained with
\begin{align}
  \begin{aligned}
    \min_{\Theta_{E_l}}\Loss_{E_l} = ||\w_{opt}-\txt||_2^2,
  \end{aligned}
\end{align}
where the $\Theta_{E_l}$ denotes the parameters of the text encoder $E_l$, which is latent space alignment module, and $\w_{opt}$ is fixed during training. 

This is a simple yet effective way to learn the alignment between text encoder $E_l$ feature space and StyleGAN2 latent space $\W$, which can leverage the disentanglability of StyleGAN2 and uncover the hidden semantic structures of latent space $\W$, such that the generated images can be manipulate-able through the text. After the training of the latent space alignment model, we input the text data into the trained $E_l$, the output text representations can be regarded as the latent codes $\w$ for StyleGAN2 to generate images. We present generation visualizations in Section \ref{qua}.

\subsection{Online paired data generation}
Based on the captions, we obtain the vocabulary $V$ consisting of all the existing words of the given datasets. Moreover, we adopt the spaCy \cite{spacy} library to give Part-of-Speech (POS) tagging, such as adj. and noun., for each word token in the captions. Then we also collect a POS vocabulary $V_{pos}$, where a set of words can be retrieved from the POS tagging.

Given a list of caption word tokens $S=\{s_1, \dots, s_N\}$, we first select part of the tokens in $S$ based on the replacement rate $r$, and then randomly pick other tokens in $V$ or $V_{pos}$ of the given dataset to replace the selected original tokens. The token list after the random replacement can be denoted as the augmented text data $S'$. We then feed the augmented text $S'$ into the trained latent space alignment module $E_l$, the output $E_l(S')$ can be used as the latent codes for StyleGAN2 to generate augmented images $I'=G(E_l(S'))$. It is notable that the replacement rate $r$ and augmentation strategy (e.g. using $V$ or $V_{pos}$) are the hyper-parameters, the full procedure is presented in Algorithm \ref{algo:1}.

In one mini-batch, we have the original text-image pairs $D=(S, I)$ and the augmented pairs $D'=(S', I')$ for training. Here we use the proposed online paired data augmentation method for the cross-modal retrieval task. $E_{txt}(\cdot)$ and $E_{img}(\cdot)$ represent text and image encoders respectively. $F_D = (E_{txt}(S), E_{img}(I))$ and $F_{D'} = (E_{txt}(S'), E_{img}(I'))$ denote the extracted feature sets of the original and augmented data in a mini-batch.

We adopt the triplet loss to learn the similarity between text and image data as follows
\begin{equation}
\begin{aligned}
\Loss_{tri} = \sum_{D,D'}\left[d_{ap}-d_{an}+m\right]_+ .
\end{aligned}
\end{equation}
where $m$ is the margin, $d_{ap}$ and $d_{an}$ denote the anchor-positive sample distances and anchor-negative distances respectively. Specifically, the form of the constructed triplet could be $<I_a, S_p, S_n>$ or $<I_a, S_p, I_n>$. That means when we take the image $I_a$ as the anchor sample, the paired text $S_p$ is used as the positive sample. We then select a text $S_n$ or image $I_n$ sample from different pairs as the negative sample. The triplets of text data as the anchor can be constructed in a similar way. The summation symbol means that we construct triplets and do the training for all the text and image instances of the mini-batch, including both original data $D$ and augmented data $D'$. 
To improve the effectiveness of training, we adopt the hard sample mining method used in \cite{hermans2017defense}.

\begin{algorithm}[t]
\caption{Paired cross-modal data augmentation strategy} 
\label{algo:1}
\hspace*{0.02in} {\bf Require:} 
The trained StyleGAN2 model $G$ and latent space alignment module $E_l$;\\
\hspace*{0.02in} {\bf Input:}
Textual caption data $S$, replacement rate $r$, whole vocabulary $V$ and POS vocabulary $V_{pos}$;
\begin{algorithmic}[1]
\State Initialize replacement strategy $\theta \in \{\text{random}, \text{pos} \}$;
\For{$S =\{s_1, \dots, s_N\} \in$ minibatch} 
    \State Give POS tagging $\{\text{pos}_{s_1}, \dots, \text{pos}_{s_N}\}$ for $S$;
    \State Randomly select part of tokens $S_r = \{s_j, \dots, s_k\}$ from $S$ to be replaced based on $r$;
    \If{$\theta$ is random}
        \State Randomly select new tokens $\{s'_j, \dots, s'_k\}$ from $V$; 
    \EndIf
    \If{$\theta$ is pos}
        \State For $s \in S_r$, select random tokens $s'$ from $V_{pos}(\text{pos}_s)$ to replace $s$;
    \EndIf
    \State Augmented captions $S' \gets \{s_i, s'_j, \dots, s'_k\}$;
    \State Augmented images $I' \gets G(E_l(S'))$;
\EndFor
\State \Return paired data (S', I')
\end{algorithmic}
\end{algorithm}

\section{Experiments}

\subsection{Dataset and experiment settings} \label{data}
We conduct experiments on two public datasets, i.e. CUB 200-2011 \cite{wah2011caltech} and Recipe1M \cite{salvador2017learning} datasets. 

\noindent \textbf{CUB.} CUB dataset consists of 11,788 images from 200 bird classes, where each image has 10 English textual descriptions. We evaluate our proposed method on CUB dataset with two types of cross-modal retrieval tasks, i.e. instance-level and class-level retrieval. For instance-level retrieval training, we only take the image and its corresponding captions as the matched pairs. We follow the official splits \cite{wah2011caltech}, where there are 5,994 and 5,794 pairs for training and test respectively.
While in the class-level retrieval, we regard all the image and text instances from the same class as the positive samples. We follow the class splits used in \cite{chun2021probabilistic}, we use 150 classes for training and validation, where 80$\%$ is used for training and 20$\%$ is used for validation. The remaining 50 classes are used for the test.

\noindent \textbf{Recipe1M.} Recipe1M is a large-scale cross-modal food dataset, collected from various cooking websites. It contains user-uploaded food image and recipe data, which includes the food titles, ingredients and cooking instructions. There are in total 238,999 training food image-recipe pairs, 51,119 and 51,303 pairs for validation and test respectively. We follow previous works \cite{salvador2017learning,wang2019learning,salvador2021revamping} to do instance-level retrieval. To be specific, we are interested in finding the corresponding recipe for the given food image query, and vice versa.

\noindent \textbf{Evaluation metrics.} We follow previous works \cite{zhang2018deep,salvador2017learning} to use Recall@\{1, 5, 10\} to do the performance evaluation. In CUB and Recipe1M instance-level retrieval, we randomly sample 1000 pairs for 10 times and report the averaged metrics results. In CUB class-level retrieval, we adopt the R-Precision (R-P) metric as an alternative to do evaluation on the whole test set. R-P is used in \cite{musgrave2020metric,chun2021probabilistic}, it computes the ratio of positive items in the top-$R$ retrieved items, where $R$ is the number of references in the same class as the query. This metric can get the best scores only if it retrieves all the positive items on the top-$R$ retrieved results.

\subsection{Implementation details} \label{implement}
We use the PyTorch implementation of StyelGAN2-ADA \cite{karras2020training} with the auto settings to train the image generation model on CUB and Recipe1M datasets, with the default training setting. We use a one-layer bi-directional LSTM as the latent space alignment module, the output dimension is identical with the dimension of the StyleGAN2 latent codes $\w$, which is $512$.

In cross-modal retrieval training, for CUB, we use a pretrained ResNet-50 \cite{he2016deep} model and a DistilBERT \cite{sanh2019distilbert} model as the image encoder and the text encoder respectively, where the output feature dimension is $512$. For Recipe1M, we implement the proposed cross-modal data augmentation method on the state-of-the-art work \cite{salvador2021revamping}. They \cite{salvador2021revamping} adopt transformers \cite{vaswani2017attention} to encode the recipes and ResNet-50 \cite{he2016deep} for images, where the output dimension is $1024$. We train the model with batch size of $32$ (resp. $128$) for CUB (resp. Recipe1M). For both datasets, we use the Adam \cite{kingma2014adam} optimizer with the initial learning rate of $0.0001$ and the learning rate decays 0.1 after 30 epochs. We set the margin $m$ as $0.3$. We use the model having the best R@1 on the validation set for testing.

We first use the augmented cross-modal data to pretrain the retrieval model for 100 (5) epochs for CUB (resp. Recipe1M) dataset. Then we finetune the pretrained backbone on the real data. We run the experiments on a single Tesla V100 GPU. The pretraining (resp. finetuning) process costs 5 hours (resp. 2 hours) and 1 day (resp. 3 days) for CUB and Recipe1M respectively.

\begin{table}
\caption{R@K (\%) of CUB instance-level retrieval models trained with \textbf{random replacement strategy}. We show model evaluation results of various replacement rates $r$, where $r=0$ denotes the baseline and $r=0.7$ achieves the best performance.}
\vspace{-10pt}
\label{tab:random}
\resizebox{0.48\textwidth}{!}{
  \begin{tabular}{l|ccccccccccc}
\toprule
$r$     & 0 & 0.1 & 0.2 & 0.3  & 0.4   & 0.5   & 0.6   & 0.7   & 0.8  & 0.9   & 1     \\
\midrule
\textbf{R@1}     & 13.0  & 13.9 & 14.2 &  15.0   & 15.3 & 14.4 & 15.3 & \textbf{15.8}           & 15.6 & 15.0 & 14.2 \\
\textbf{R@5} & 34.8  & 37.5 & 39.6 & 39.2 & 40.9  & 40.2  & 40.3  & \textbf{41.1}  & 40.1  & 40.0  & 39.1  \\
\textbf{R@10}   & 49.5  & 50.5 & 52.6 & 52.3 & 52.6 & 52.8 & 52.3 & \textbf{54.0} & 53.3 & 51.5  & 51.3 \\
\bottomrule
  \end{tabular}}
\vspace{-10pt}
\end{table}

\subsection{Results of different replacement strategies} \label{aug}
Here we choose to use CUB instance-level image-to-text retrieval to show the results of the ablation studies. In CUB, the unpaired caption and image data of the same class, can vary due to the differences on the bird poses, shapes and colours, hence the instance-level retrieval is more challenging than the class-level one and can reflect more fine-grained information from the model. We take the same training method across all the experiments, where we first train the model with the adopted data augmentation techniques, then we finetune the trained model with the original training set. 

We have two text replacement strategies, i.e. random replacement and POS replacement. The difference between the two strategies is that, random replacement randomly selects other words from the whole vocabulary to replace, while POS replacement considers the POS tagging of the replaced word and retrieve another random word with the same POS tagging to replace.

\begin{table}
\caption{R@K (\%) of CUB instance-level retrieval models trained with \textbf{POS replacement strategy}. The headings indicate the tokens with certain POS tagging are replaced. Using the random replacement strategy yields the best results.}
\vspace{-10pt}
\label{tab:pos}
\resizebox{0.42\textwidth}{!}{
\begin{tabular}{l|c|ccccc}
\toprule
$r=0.7$ &\textbf{random} &\textbf{all tagging} &\textbf{adj.} &\textbf{noun.} &\textbf{adj.+nonn.} \\\midrule
\textbf{R@1} & \textbf{15.8} &15.3 &14.6 &14.8 &14.9 \\
\textbf{R@5} & \textbf{41.1} &39.6 &40.1 &39.5 &39.8 \\
\textbf{R@10} & \textbf{54.0}  &51.9 &52.3 &52.5 &52.0 \\
\bottomrule
\end{tabular}}
\vspace{-10pt}
\end{table}

\noindent\textbf{Random replacement strategy.} In Table \ref{tab:random}, we show the evaluation performance of models trained with random replacement strategy. To be specific, we experiment with various replacement rates $r$ to construct the augmented data, $r=0$ denotes the baseline performance, where we do not perform the proposed augmentation method.
We observe $r=0.7$ yields the best performance, it can improve the baseline performance by $21.5\%$ and $9.1\%$ in R@1 and R@10 respectively, which demonstrates the effectiveness of the augmented cross-modal data for metric learning. When $r=1$, although the generated text would be completely random, it still contains some semantic words, hence we can see $r=1$ also outperforms the baseline model ($r=0$) without using the augmented data. This suggests our method can increase the diversity of the paired data and produce more meaningful text-image pairs compared to the traditional single-modality data augmentation methods. 

\noindent\textbf{POS replacement strategy.} In Table \ref{tab:pos}, we present results of the POS replacement strategy, where the replacement rate $r=0.7$ is used for all the following experiments. We can see the results also outperform the baseline. However, there is a performance gap between POS and random replacement, where the model trained with POS replacement has over $2\%$ drop on R@10 compared to that trained with random replacement. The reason can be when we select new word tokens from the vocabulary of certain POS tagging only, the diversity is decreased. It might suggest the importance of the diverse combinations of the cross-modal data for retrieval training.

\begin{table}
 \caption{R@K (\%) of CUB instance-level retrieval models trained with different settings.}
 \vspace{-10pt}
 \label{tab:ablation}
 \centering
 \resizebox{0.4\textwidth}{!}{
 \begin{tabular}{l|c|c|c}
\toprule
\textbf{Method}      & \textbf{R@1} & \textbf{R@5} & \textbf{R@10} \\
\midrule
baseline ($r=0$) & 13.0      & 34.8    & 49.5                     \\
best ($r=0.7$) & \textbf{15.8}     & 41.1   & 54.0        \\
\midrule
\midrule
\multicolumn{4}{c}{\textbf{Effect of augmented data scale}}              \\
\midrule
90\% of real data & 15.7 & \textbf{42.1} & \textbf{54.6} \\
110\% of real data & 15.2 & 40.8 & 53.4 \\
\midrule
\midrule
\multicolumn{4}{c}{\textbf{Text-conditioned StyleGAN2}}              \\
\midrule
$r=0.7$              & 14.6  & 40.4 & 52.9                     \\
\midrule
\midrule
\multicolumn{4}{c}{\textbf{Augmented data as noise}}              \\
\midrule
$r=0.2$              & 13.9   & 37.7 & 49.7                     \\
$r=0.3$              & 14.2   & 37.4 & 49.9                     \\
\midrule
\midrule
\multicolumn{4}{c}{\textbf{LSTM as text encoder}}              \\
\midrule
no augmentation & 5.4 & 18.0 & 26.7\\
augmentation & 7.5 & 22.0 & 31.5 \\
\midrule
\midrule
\multicolumn{4}{c}{\textbf{Single-modality data augmentation}}              \\
\midrule
augmented text only    & 14.2                    & 38.3                    & 50.0 \\ 
unpaired image and text & 14.3                    & 39.0                    & 51.4                     \\
\midrule
\midrule
\multicolumn{4}{c}{\textbf{Prevailing data augmentation methods}}              \\
\midrule
VSE$\infty$ \cite{chen2021learning} & 13.7 & 36.9 & 49.1 \\
Rand-Augment & 14.2 &	40.2 &	52.3	\\
RandomErasing & 14.0 & 38.6 & 51.3  \\
Rand-Augment + RandomErasing & 14.8 & 38.3 & 50.7 \\
\bottomrule
\end{tabular}}
\vspace{-15pt}
\end{table}

\subsection{Ablation studies} 

We further demonstrate results of more augmentation settings in Table \ref{tab:ablation}. 

\noindent\textbf{Effect of different augmented data scale.} Our default setting is to do the online augmentation, which means the augmented data has the same size as the real data. Here we keep $r=0.7$ and set the augmented dataset size as 90\% and 110\% of the real data. We observe scaling the augmented data size as 90\% of the real data size gets the best results on R@5 and R@10.

\noindent\textbf{Using text-conditioned StyleGAN2 for augmentation.} We use the CLIP \cite{radford2021learning} model to give text embeddings and implement the conditional StyleGAN2, where we directly input text into StyleGAN2 and obtain the generated images. Specifically, we observe the FID score of conditional generation is \textbf{5.52}, while the FID of our proposed method is \textbf{4.54} (lower is better). It indicates our generated images have better quality. The quantitative results of Table \ref{tab:ablation} also show our method has better performance on cross-modal data augmentation than text-conditioned StyleGAN2. We give more discussions on Section \ref{dis}.

\noindent\textbf{Using augmented data as noise.} We further experiment with small replacement rate $r \le 0.3$ in Table \ref{tab:ablation}, where we assume the semantic contents of augmented data $D'$ only has slight change compared with the original data $D$. Hence we set $D'$ to remain as the same pair with $D$. 
In this case, $D'$ can be regarded as the noisy image-text pairs, as they are not totally matched with $D$. We can see that this way also boosts the model performance, since adding the noisy pairs to the training process enables the retrieval model to be more noise-resistant.

\noindent\textbf{Results of LSTM as the text encoder.} When we change the text encoder from the pretrained DistilBERT to a plain LSTM, we can see the model performance drops heavily, where the R@10 score decreases over $46\%$. However, the model trained with our proposed method can still improve the LSTM baseline performance across all the evaluation metrics. It suggests the generalization of our algorithm.

\noindent\textbf{Single-modality data augmentation.} To evaluate if the latent space alignment module can give meaningful pair information, we run the experiments with the augmented text $S'$ only, we also assign unpaired labels to the augmented text $S'$ and image $I'$ and train the model. In this experiment, we use the random replacement strategy and $r$ is set as $0.7$, which are the best setting in Table \ref{tab:random}. It can be seen from Table \ref{tab:ablation}, when we adopt $S'$ or unpaired $S'$ and $I'$ for training, both methods yield inferior performance compared to the models trained with the paired augmented data.

\noindent\textbf{Results of using prevailing data augmentation methods.} VSE$\infty$ \cite{chen2021learning} propose to mask partial input data as the data augmentation strategy, where they do not produce paired augmented image-text data but use masked data only, hence they have inferior results than ours. We also follow \cite{touvron2021training} to implement two prevailing data augmentation methods for images, i.e. Rand-Augment \cite{cubuk2020randaugment} and RandomErasing \cite{zhong2020random}. We set $N=1, M=2$ and erasing probability $p=0.25$ in Rand-Augment and RandomErasing respectively, which are the empirical settings in \cite{touvron2021training}. We use the augmented data to pretrain the model, and then finetune the model without the Rand-Augment or RandomErasing augmented data. We observe these image data augmentation methods both can improve the baseline performance, while they still have performance gap with our proposed paired data augmentation method.

\begin{table}
 \caption{Evaluation of the performance of our proposed method compared against various CUB class-level retrieval benchmarks. The models are evaluated with R-Precision (\%) and R@K (\%). In baseline model, we use the pretrained ResNet-50 and DistilBERT without our augmented data.}
 \vspace{-10pt}
 \label{tab:cub}
 \centering
 \resizebox{0.35\textwidth}{!}{
 \begin{tabular}{l|cc|cc}
\toprule
\multicolumn{1}{l|}{\multirow{2}{*}{\textbf{Method}}}           & \multicolumn{2}{c|}{\textbf{Image-to-text}} & \multicolumn{2}{c}{\textbf{Text-to-image}} \\
\multicolumn{1}{c|}{}                                  & R-P             & R@1             & R-P             & R@1             \\
\midrule
VSE0 \cite{faghri2017vse++}         & 22.4            & 44.2            & 22.6            & 32.7            \\
PVSE \cite{song2019polysemous}       & 18.4            & 47.8            & 19.9            & 34.4            \\
PCME    \cite{chun2021probabilistic} & 26.3            & 46.9            & 26.8            & 35.2            \\
\quad + Ours  &      27.3       &  48.4         &   27.6         & 37.3 \\
\midrule
baseline                                              & 27.4            & 46.4            & 27.9           & 37.2            \\
\quad + Ours  & \textbf{30.0}            & \textbf{52.7}            & \textbf{29.7}            & \textbf{40.6} \\
\bottomrule
\end{tabular}}
\vspace{-15pt}
\end{table}

\begin{table*}
\centering
  \caption{Evaluation of the performance of our proposed method compared against various Recipe1M instacne-level retrieval benchmarks. The models are evaluated on the basis of R@K (\%).}
  \vspace{-10pt}
  \label{tab:food}
\resizebox{0.8\textwidth}{!}{
\begin{tabular}{l|ccc|ccc|ccc|ccc}
\toprule
\multicolumn{1}{c|}{\multirow{3}{*}{}} & \multicolumn{6}{c|}{\textbf{1k}}                       & \multicolumn{6}{c}{\textbf{10k}}               \\
\cmidrule{2-13}
\textbf{Method} & \multicolumn{3}{c|}{\textbf{image-to-recipe}} & \multicolumn{3}{c|}{\textbf{recipe-to-image}} & \multicolumn{3}{c|}{\textbf{image-to-recipe}}  & \multicolumn{3}{c}{\textbf{recipe-to-image}} \\
& R@1        & R@5       & R@10       & R@1        & R@5       & R@10       & R@1        & R@5       & R@10       & R@1        & R@5       & R@10         \\
\midrule
Salvador et al. \cite{salvador2017learning}    & 24.0   & 51.0   & 65.0   & 25.0   & 52.0 & 65.0   & -    & -    & -    & -    & -    & -           \\
Chen et al. \cite{chen2017cross} & 25.6 & 53.7 & 66.9 & 25.7 & 53.9 & 67.1 & 7.2  & 19.2 & 27.6 & 7.0    & 19.4 & 27.8      \\
Carvalho et al. \cite{carvalho2018cross} & 39.8 & 69.0   & 77.4 & 40.2 & 68.1 & 78.7 & 14.9 & 35.3 & 45.2 & 14.8 & 34.6 & 46.1          \\
R2GAN \cite{zhu2019r2gan}    & 39.1 & 71.0   & 81.7 & 40.6 & 72.6 & 83.3 & 13.5 & 33.5 & 44.9 & 14.2 & 35.0   & 46.8         \\
MCEN \cite{fu2020mcen}         &     48.2           & 75.8          & 83.6 & 48.4          & 76.1          & 83.7          & 20.3          & 43.3          & 54.4          & 21.4          & 44.3          & 55.2         \\
ACME \cite{wang2019learning}      & 51.8           & 80.2          & 87.5 & 52.8          & 80.2          & 87.6          & 22.9          & 46.8          & 57.9          & 24.4          & 47.9          & 59.0         \\
SCAN \cite{wang2021cross}   & 54.0            & 81.7          & 88.8 & 54.9          & 81.9          & 89.0    & 23.7          & 49.3          & 60.6          & 25.3          & 50.6          & 61.6          \\
DaC \cite{fain2019dividing}  & 55.9           & 82.4          & 88.7 & -             & -             & -             & 26.5          & 51.8          & 62.6          & -             & -             & -           \\
\midrule
SOTA \cite{salvador2021revamping}    &    59.1 & 86.9 & 92.3 & 59.1 & 87.0   & 92.7 & 27.3 & 55.4 & 67.3 & 27.8 & 55.6 & 67.3         \\
\quad + Ours                 & \textbf{60.6} & \textbf{87.7} & \textbf{92.8} & \textbf{61.3} & \textbf{87.7} & \textbf{93.2} & \textbf{28.6} & \textbf{57.1} & \textbf{68.6} & \textbf{29.3} & \textbf{57.3} & \textbf{69.0} \\
\bottomrule
\end{tabular}}
\vspace{-5pt}
\end{table*}

\subsection{Results against existing methods}
Here we take the augmentation settings with the best validated performance in Section \ref{aug}, where we adopt $r=0.7$ and the random text replacement strategy, to do the following experiments.

The existing cross-modal retrieval works on CUB \cite{faghri2017vse++,song2019polysemous,chun2021probabilistic} mainly focus on the class-level retrieval setting, we compare our proposed method against various models in Table \ref{tab:cub}, where the results of \cite{faghri2017vse++,song2019polysemous,chun2021probabilistic} are taken from \cite{chun2021probabilistic}. We follow the prevailing batch construction method \cite{faghri2017vse++,lee2018stacked,li2019visual,chun2021probabilistic}, where the positive and the negative samples are from the different modality of the anchor samples.
We train our baseline model by the triplet loss, where we use a pretrained ResNet-50 and DistilBERT model as image and text encoders respectively. 
We add the proposed paired data augmentation method on PCME \cite{chun2021probabilistic} and our baseline model, to demonstrate the efficacy of our method.
The model trained with our augmented data boosts the baseline performance across all the metrics. The augmented baseline model improves $9\%$ and $14\%$ on image-to-text R-Precision and R@1 scores respectively. 

We show results on Recipe1M in Table \ref{tab:food}. Recipe1M contains large amounts of recipe data and food images in the wild, hence it is more challenging than the CUB dataset. Existing methods mainly focus on how to improve the aligned feature representations of images and recipes. For example, Wang et al. \cite{wang2019learning} adopt the adversarial training and cycle consistency to learn the cross-modal embeddings. The current state-of-the-art (SOTA) method \cite{salvador2021revamping} introduce to use transformer as the recipe encoder, they apply the triplet loss not only on the paired images and recipes, but also on the recipe components, so that the semantic relationships within recipes can be leveraged. It can be seen that our augmented model further improves the performance of SOTA in the 1k and 10k retrieval settings, yielding the best results across all the metrics. Specifically, given that Recipe1M already provides abundant data with large diversity, the improvements over the SOTA at Recipe1M demonstrates the efficacy of our proposed data augmentation algorithm on the large and complex cross-modal datasets.

\begin{table}
\centering
\caption{The evaluation on the test split of the COCO dataset.}
\vspace{-10pt}
\label{tab: coco}
\resizebox{0.48\textwidth}{!}{
\begin{tabular}{l|r|rrr|rrrr}\toprule
&\multirow{2}{*}{\textbf{Method}} &\multicolumn{3}{c|}{\textbf{image-to-text}} &\multicolumn{3}{c}{\textbf{text-to-image}} \\
& &R@1 &R@5 &R@10 &R@1 &R@5 &R@10 \\\midrule
\multirow{2}{*}{ResNet-50 \cite{he2016deep}} &baseline &49.2 &80.1 &89.7 &43.4 &79.6 &89.9 \\
&\quad + Ours  &49.9 &80.4 &89.8 &44.5 &79.6 &89.5 \\
\midrule
\multirow{2}{*}{Beit \cite{bao2021beit}} &baseline &58.6 &84.7 &91.4 &47.3 &81.9 & \textbf{90.8} \\
&\quad + Ours  & \textbf{60.2} & \textbf{85.3} & \textbf{92.3} & \textbf{49.0} & \textbf{82.1} &90.3 \\
\bottomrule
\end{tabular}}
\vspace{-10pt}
\end{table}

\subsection{Results on COCO dataset} 
Our proposed paired data augmentation framework is mainly applicable to image-text datasets where image data has single objects, as the StyleGAN2 model can hardly produce high-quality images containing complex objects and scenarios. Nevertheless, here we show the experiment results on the COCO \cite{lin2014microsoft} dataset to see the efficacy of our proposed method on the dataset with complex scenarios.

The COCO dataset contain 123,287 images in total, where each image has 5 captions. The dataset is split into 113,287 images for training, 5,000 images for validation and 5,000 images for testing. We follow previous evaluation practice \cite{Diao2021SGRAF,lee2018stacked}, and report results of the average over 5 folds of 1K test images.

It is notable that most existing state-of-the-art cross-modal retrieval frameworks \cite{li2020oscar,kim2021vilt,Diao2021SGRAF,lee2018stacked} on COCO are based on the extracted object features from images, while we generate the augmented images online, it is unrealistic to adopt the object detection model to online extract the object features from the augmented images with limited computation resources. Hence we use a relatively simple end-to-end training framework to conduct COCO experiments, which is the same as the structure in Figure \ref{fig:framework}.

Specifically, we experiment with two types of image encoders, i.e. the pretrained ResNet-50 \cite{he2016deep} and the vision transformer Beit \cite{bao2021beit}. We adopt the pretrained DistilBERT \cite{sanh2019distilbert} model as the text encoder. We present the evaluation results in Table \ref{tab: coco}. It is observed that our proposed paired data augmentation method boosts the baseline performance greatly at the evaluation metrics of R@1, while may not have great performance improvement at R@10. It might indicate that the augmented data helps more on identifying the top-ranked retrieved samples, since the data augmentation gives more data variety for the training process.

\begin{figure*}
\begin{center}
\includegraphics[width=0.95\textwidth]{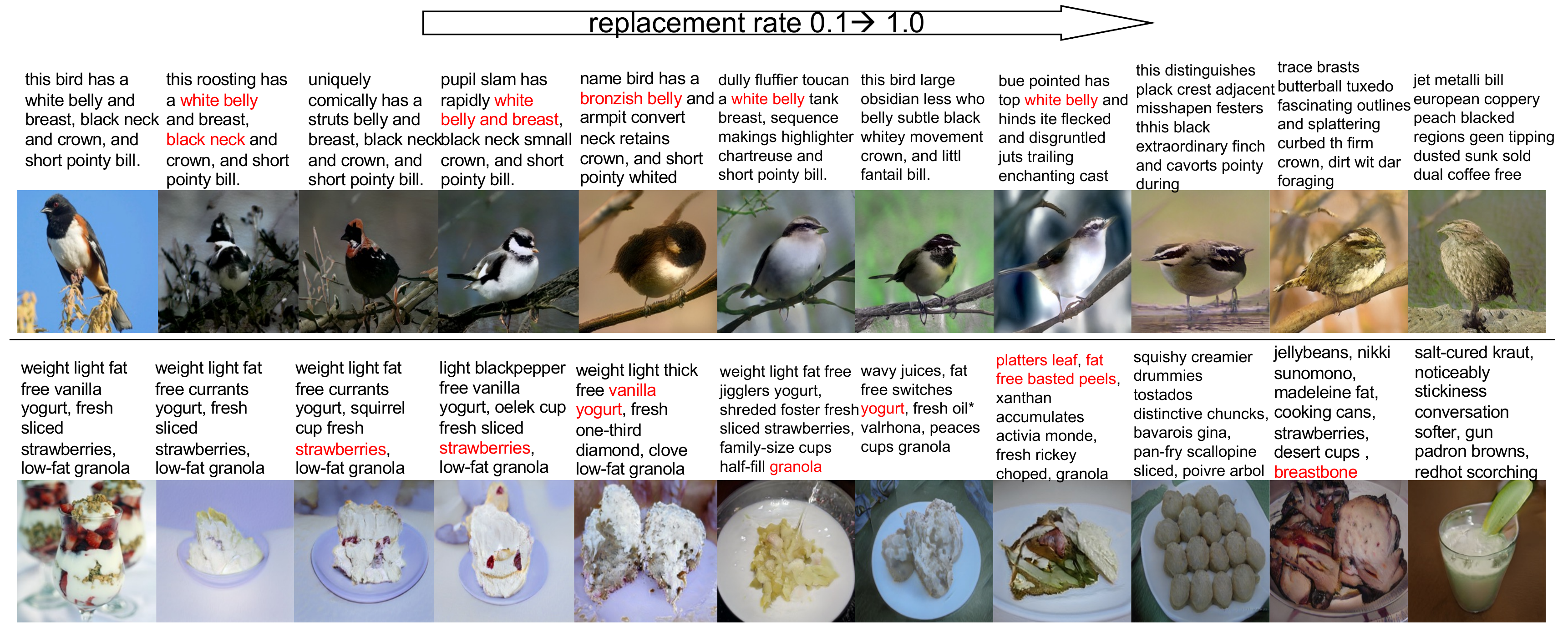}
\end{center}
\vspace{-10pt}
   \caption{The visualizations of the augmented images generated from the augmented text, where we use the random replacement strategy. From left to right, we show the original text and images, and then the augmented data produced with various replacement rates from 0.1 to 1. The words in red colour denote the correspondence with the generated images.
}
\vspace{-5pt}
\label{fig:vis}
\end{figure*}

\begin{figure}
\begin{center}
\includegraphics[width=0.48\textwidth]{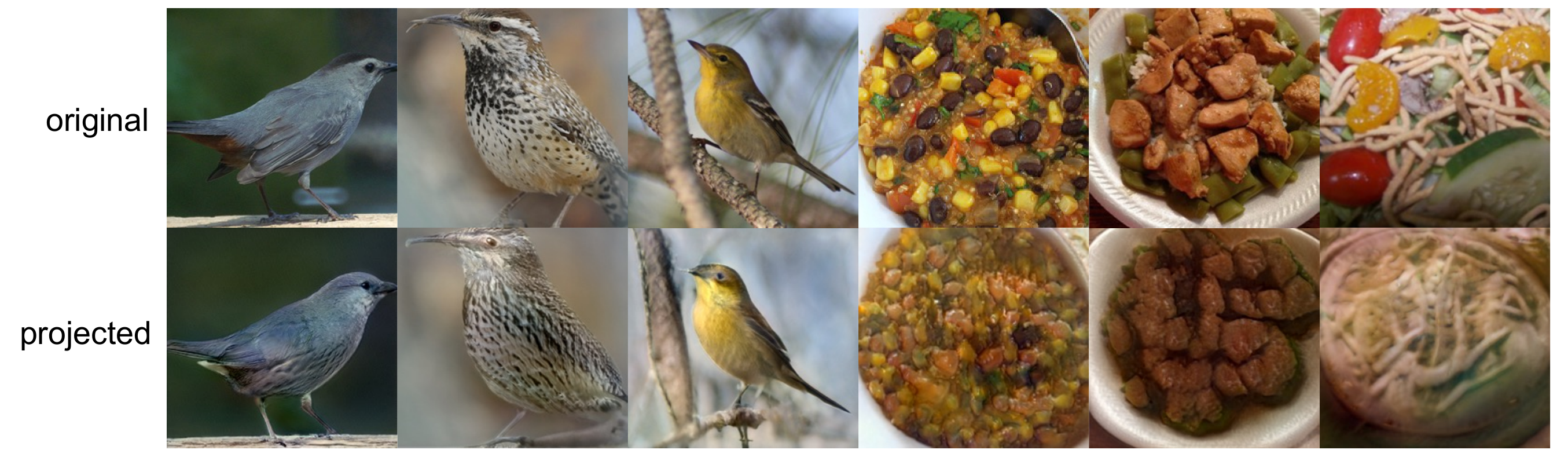}
\end{center}
\vspace{-10pt}
   \caption{The original and projected images.
}
\vspace{-10pt}
\label{fig:pro}
\end{figure}

\subsection{Qualitative results} \label{qua}
We show the qualitative results of paired augmented cross-modal data in Figure \ref{fig:vis}. It is notable that, in Recipe1M dataset the text (recipe) data includes the food titles, ingredients and cooking instructions, we do random replacement on all the components of recipes. Here we only show the augmented ingredients in Figure \ref{fig:vis} for simplicity. Besides, some word typos can be seen in the visualizations, which are actually contained in the original vocabulary of the given dataset.

To be specific, from left to right in Figure \ref{fig:vis}, we present the original data and the augmented data generated from the replacement rates $r$ from 0.1 to 1. We can observe that the StyleGAN2 generated images based on the output of the latent space alignment module, can generally produce images with decent quality and enough diversity. Although the augmented text with the random replacement transforms looks plausible and meaningless, we can still find some correspondence between the augmented text and the generated images. For example, in the first row, when we set $r$ as 0.3, the augmented text contain the semantic information of \emph{white belly and breast} and \emph{black neck and crown}, we can see the text semantics has been reflected in the generated image. When we change $r$ to 0.4, where the text has \emph{bronzish belly}, the bird color of the augmented image can also change accordingly. In the second row, the augmented data has minor changes within $r \in [0.1, 0.3]$, as the length of each ingredient is short. When $r$ increases, we can see the large diversity of the augmented images.

Moreover, we show the comparison between the original and projected images in Figure \ref{fig:pro}. The projected images are generated from the projected $\w$ of Equation \ref{eq:pro}. We observe most projected CUB images can preserve the original image semantic information and object textures. While in Recipe1M dataset, where the food images contain more fine-grained ingredients, the projected images may fail to preserve all the original information, since the StyleGAN2 model is not capable of generating perfect images for such complex scenarios. Nevertheless, our proposed algorithm can produce large amounts of plausible images with diverse semantics and improve the retrieval performance.

\section{Discussion} \label{dis}
\noindent \textbf{Why to use StyleGAN instead of the text-conditioned GAN?} When we train a generation model using the text-conditioned GAN structure, we only have limited text-image pairs, thus the model can hardly produce diverse images compared to the unconditioned GAN model. The recent proposed StyleGAN gives a new perspective to generate manipulate-able images, to be specific, it introduces to map the random noise $\z$ to another latent codes $\w$, and the images are generated from the mapped latent codes $\w$. In this way, the StyleGAN model can learn a more disentangled latent space $\W$. Some research works \cite{xia2021tedigan,wang2021cycle} demonstrate this architecture outperforms the conditional GAN on the text-to-image generation task. Therefore, here we take the advantage of the disentanglability of StyleGAN by uncovering the semantics of the latent space $\W$ and further manipulate the augmented images through the latent space $\W$. Our experiment results also depict its superiority over the text-conditioned GAN on the cross-modal data augmentation task, as shown in Table \ref{tab:ablation}.

\section{Conclusion}
This paper proposed a novel framework for paired cross-modal data augmentation, which can generate an unlimited amount of paired data for training cross-modal retrieval models. Specifically, we use the strategy of random text replacement to produce the augmented text. To generate the corresponding augmented images from the augmented text, we first adopt the StyleGAN2 model to generate images with high quality and diversity. Then we proposed to bridge the gap between the text and image data by our latent space alignment module, which maps the text features into the latent space $\W$ of StyleGAN2. We use the output of the learned alignment module for StyleGAN2 to generate the augmented images, thus allow us to obtain the augmented text-image pairs. We further evaluate the quality of the augmented data, through the image-to-text retrieval task on two public datasets. The promising experimental results demonstrate our proposed method can effectively boost the performance of state-of-the-art models on several different benchmarks.

\section*{Acknowledgments}
This research is supported, in part, by the National Research Foundation (NRF), Singapore under its AI Singapore Programme (AISG Award No: AISG-GC-2019-003) and under its NRF Investigatorship Programme (NRFI Award No. NRF-NRFI05-2019-0002). Any opinions, findings and conclusions or recommendations expressed in this material are those of the authors and do not reflect the views of National Research Foundation, Singapore. This research is supported, in part, by the Singapore Ministry of Health under its National Innovation Challenge on Active and Confident Ageing (NIC Project No. MOH/NIC/HAIG03/2017).
This research is also supported by the National Research Foundation, Singapore under its AI Singapore Programme (AISG Award No: AISG-RP-2018-003), and the MOE AcRF Tier-1 research grant: RG95/20.

\bibliographystyle{ACM-Reference-Format}
\bibliography{sample-base}

\end{document}